\pdfoutput=1

\documentclass[11pt]{article}

\usepackage{graphicx}  % \usepackage[dvipdfmx]{graphicx}  % \usepackage{bmpsize}
\usepackage{booktabs}
\usepackage{amsmath,amssymb,bm,stmaryrd,bbm}
\usepackage{mathtools}
\usepackage{titling}
\usepackage{authblk}
\usepackage{xcolor}
\usepackage{fullpage}
\usepackage[symbol]{footmisc}

\newcommand{\appendixcontext}[2]{#1}

\DeclarePairedDelimiter{\ceil}{\lceil}{\rceil}

\makeatletter
\def\input@path{{./tex/}}
\makeatother

\title{Federated Survival Analysis with Discrete-Time Cox Models}
\author[1$\S$]{Mathieu Andreux\thanks{Corresponding author: \texttt{mathieu.andreux@owkin.com}.}}
\author[2$\dagger\S$]{Andre Manoel}
\author[1$\S$]{Romuald Menuet}
\author[1$\S$]{\authorcr Charlie Saillard}
\author[3$\dagger\S$]{Chlo\'{e} Simpson}
\affil[1]{Owkin Inc., New York, USA.}
\affil[2]{Hospital Israelita Albert Einstein, S\~{a}o Paulo, Brazil}
\affil[3]{French Red Cross Innovation, Montrouge, France}

\begin{document}
\maketitle

\footnotetext[2]{Work done while at Owkin.}
\footnotetext[4]{Alphabetical order.}

\begin{abstract}
Building machine learning models from decentralized datasets 
located in different centers with 
federated learning (FL) is a promising approach to circumvent local data 
scarcity while preserving privacy.
However, the prominent Cox proportional hazards (PH) model, used for
survival analysis, does not fit the FL framework, as its loss function is non-separable
with respect to the samples.
The na\"{i}ve method to bypass this non-separability consists 
in calculating the losses per center, and minimizing their sum as an 
approximation of the true loss. 
We show that the resulting model may suffer from important performance 
loss in some adverse settings.
Instead, 
we leverage the discrete-time extension
of the Cox PH model to formulate survival analysis 
as a classification problem with a separable loss function.
Using this approach, we train survival models 
using standard FL techniques on
synthetic data, as well as real-world datasets from The Cancer Genome Atlas (TCGA),
showing similar performance to a Cox PH model trained on aggregated data.
Compared to 
previous works, the proposed method is more communication-efficient, 
more generic, and more amenable to using privacy-preserving techniques. 
\end{abstract}

\section{Introduction}
\label{introduction}

The goal of survival analysis is to infer the occurrence time of some event
for individuals, be it failure times in the case of industrial
installations, or adverse events for patients in a clinical
setting~\cite{kleinbaum2010survival}.  In the latter case, the resulting
models, often based on the Cox proportional hazards (PH)
model~\cite{cox1972regression}, are routinely used by clinicians for
prognostic purposes and by biomedical researchers for understanding
diseases, see \textit{e.g.}~\cite{wilson2012international}
and~\cite{courtiol2019deep} respectively.

One of the characteristics of clinical data is its relative global
abundance, while being scarce in individual centers, especially for rare
diseases.  Aggregating local datasets is often impossible due to their
sensitivity and strict privacy regulations.  In this
setting, federated learning~\cite{shokri2015privacy,
mcmahan2017communication} is a promising approach as it enables
collaboratively training models over multiple centers without moving the
datasets.

This paper investigates the problem of training Cox PH survival models in this
federated setting. This setting is challenging since the Cox PH loss
function is not separable with respect to either centers or individual
records, and therefore does not fit into existing frameworks for federated
learning.

Among existing works tackling this problem, the WebDISCO
algorithm~\cite{lu2015webdisco} is the closest to our setting.  While it
permits one to train the Cox PH model on decentralized datasets, it has a
large communication cost, is limited to linear models, and may expose some
private data, as we show in Sec.~\ref{previous}. In this paper, we target a
communication-efficient, generic, and privacy-preserving method for federated
survival analysis.

\subsection{Main contributions}
\begin{enumerate}
    \item We show that na\"{i}vely doing federated learning with the
        non-separable Cox loss leads to the so-called stratified Cox model,
        with per-center stratification (Sec.~\ref{motiv}).  Moreover, we
        demonstrate how, in extreme cases, this stratification might have
        severe consequences for the model's predictive performance (Sec.
        \ref{sec:synthetic}).
    \item We provide a workaround to this issue by approximating the Cox
        loss with a discrete-time model (Sec.~\ref{methods}) which,
        asymptotically, provides the same results as the original
        continuous-time model~\cite{zhong2019survival}. Since the loss for
        this model is separable, it can be easily plugged in existing
        federated learning algorithms.  We empirically study the impact of
        this approximation in both synthetic and real-world tabular datasets
        (Sec.~\ref{sec:synthetic} and~\ref{sec:realworld}).
    \item We use the proposed method to analyze multimodal datasets from The
        Cancer Genome Atlas (TCGA), which provide both image and tabular
        patient data along with survival statistics. We demonstrate the
        flexibility of our approach by combining it with deep learning
        models and training it on large whole-slide histopathology images
        (Sec.~\ref{sec:histo}).
\end{enumerate}

\section{Background}
\label{background}

\subsection{Survival analysis}
\label{back_surv}

Given a group of individuals susceptible to experiencing an event of
interest, survival analysis techniques seek to infer the probability
distribution for the time of that event for each
individual~\cite{kleinbaum2010survival}. Such techniques are widely used
in medical settings, \textit{e.g.} for prognostic purposes,
where individuals are patients and the event can be the onset of a
disease, admission to a hospital, or death.

Obtaining these times-to-event requires patients to be followed in long
observational studies, which typically last years. Due to this long
duration, patients frequently drop out from studies before any event occurs.
In this case, one can only know that the event did not occur before the
dropping time, which is known. This partial knowledge is known in survival
analysis as \emph{right censoring}.

\paragraph{Notation} 
We consider individuals $i \in [N] = \{1, \dots, N\}$, each represented by a
tuple $(\bm{x}_i, t_i, \delta_i)$: a vector of covariates $\bm{x}_i \in
\mathbb{R}^P$, an observed time point $t_i \in \mathbb{R}_+$, and an
indicator of whether or not an event occured, $\delta_i \in \{0, 1\}$, where
$\delta_i = 0$ if the event has been censored. Due to censoring, the
observed time $t_i$ is only a lower-bound on the actual event time~$\tau_i
\in \mathbb{R}_+$ for individual $i$; these values match when~$\delta_i =
1$.

The goal of survival analysis is to model the probability distribution of
the random variable $\tau_i$ for each individual~$i$. A key quantity
characterizing this distribution is the hazard function~$\lambda(t, x_i)$,
which is the instantaneous rate of occurrence of the event given that it has
not yet happened
\begin{equation}
    \lambda (t, \bm{x}_i) = \lim_{\Delta t \to 0} \frac{\mathbb{P}\left[ t \leq \tau_i
    \leq t + \Delta t | \bm{x}_i, \tau_i \geq t \right]}{\Delta t}.
\end{equation}

\paragraph{The Cox proportional hazards model}
The Cox proportional hazards (PH) model~\cite{cox1972regression} is one of
the most used models in survival analysis. It assumes that the hazard
function can be factored as
\begin{equation}
    \lambda(t, \bm{x}_i) = \lambda_0(t) \exp(\bm{\beta}^T \bm{x}_i),
    \label{eq:coxhazard}
\end{equation}
where $\lambda_0(t)$ is the baseline hazard function---common to all
individuals in the group and dependent on time only---and $\bm{\beta}$ is
the vector of parameters of the model. Notably, the model implies that the
risk ratio between individuals does not depend on time. Therefore, when
only interested in relative comparisons, the baseline hazard function does
not need to be specified.

The function $\lambda_0 (t)$ and the parameters $\bm{\beta}$ can be
estimated independently from each other. For the latter, one can minimize
the negative Cox partial log-likelihood, which only compares relative risk
ratios and is defined by
\begin{align}
    \mathcal{L}(\bm{\beta}) = -\sum_{i:\delta_i=1} \bigg\{
    \bm{\beta}^T \bm{x}_i - \log \! \sum_{j:t_j\geq
    t_i} e^{\bm{\beta}^T \bm{x}_j} \bigg\}.
\label{eq:cox}
\end{align}
Crucially, due to the log-sum-exp term in the r.h.s. of~\eqref{eq:cox},
which depends on a ranked subset of individuals, this partial log-likelihood
cannot be written as a sum of terms depending each on a single $\bm{x}_i$.
For that reason, we say that the Cox partial log-likelihood is
\emph{non-separable}.

\paragraph{Stratification}
The stratified Cox model \cite{holt1974survival} is a variation of the Cox
PH model for a multi-center setting, in which each center can have a
different baseline hazard function but the parameters $\bm{\beta}$ are
shared across centers. It can be useful if the suspicion exists that the PH
assumption holds locally, but not globally.
As noted by
\cite{glidden2004modelling}, the resulting stratified log-likelihood then
reads as the sum of the per-center Cox log-likelihoods~\eqref{eq:cox},
\textit{i.e.}
\begin{align}
    \mathcal{L}_{\text{strat}}(\bm{\beta}) = -\sum_{k = 1}^C
    \sum_{\substack{i \in I_k \\ \delta_i=1}}
    \bigg\{ \bm{\beta}^T\bm{x}_i - \log \sum_{\substack{j \in I_k \\ t_j\geq
    t_i}} e^{\bm{\beta}^T\bm{x}_j} \bigg\},
    \label{eq:coxstratlog}
\end{align}
where $C$ is the number of centers, indexed by $k=1, \ldots, C$, and
$I_1, \ldots, I_C \subset [N]$ denote the partition of individuals
across these centers.
Crucially, the stratified Cox log-likelihood is separable at the center
level, but not at the patient level.

\paragraph{Discrete-time extension}
When the number of individual times-to-event is small, be it after
quantization or because of the data distribution, it is often profitable to
take into account the discrete nature of the observations for modelling
purposes~\cite{rodriguez2016lecture}.

In this case, let~$\left\lbrace s^{(m)} \right\rbrace_{m \in [T]}$ denote
the ordered set of unique times-to-event of size~$T$, such that~$s^{(1)} <
\ldots < s^{(T)}$. The hazard function is thus defined as a weighted sum of
Dirac functions~$\delta(t - s^{(m)})$ located at times~$s^{(m)}$,
\begin{equation}
\lambda(t, \bm{x}_i) = \sum_{m=1}^{T} p^{(m)}(\bm{x}_i) \delta(t - s^{(m)}),
\end{equation}
where $p^{(m)}(\bm{x}_i)$ is the conditional probability of individual~$i$
having an event at time~$s^{(m)}$, knowing that the individual is still at
risk at that time:
\begin{equation}
    p^{(m)}(\bm{x}_i) = \mathbb{P} \big[\tau_i = s^{(m)} \mid \bm{x}_i,
    \tau_i \geq s^{(m)} \big].
\end{equation}
The Cox PH model~\eqref{eq:coxhazard} then does not apply, due to the hazard
function being bounded in $[0, 1]$. \cite{cox1972regression}~proposes to
adapt it in the discrete-time setting as
\begin{align}
     \frac{p^{(m)}(\bm{x}_i)}{ 1 - p^{(m)}(\bm{x}_i)} =
    \frac{p_0^{(m)}}{ 1 -
    p_0^{(m)}}
    \exp \left(\bm{\beta}^T\bm{x_i}\right),
    \label{eq:coxdt}
\end{align}
with $\lbrace p_0^{(m)} \rbrace_{m \in [T]}$ leading to a baseline hazard
function. Denoting the sigmoid function by $\sigma(x) = (1 +e^{-x})^{-1}$
and introducing coefficients
$\alpha^{(m)} \triangleq \sigma^{-1}
\big(p_0^{(m)}\big)$,
\eqref{eq:coxdt} implies that the conditional survival probability
follows a logistic model with time-dependent biases~$\alpha^{(m)}$ and
time-independent weights~$\bm{\beta}$:
\begin{equation}
p^{(m)}(\bm{x}_i)= \sigma \left( \alpha^{(m)} + \bm{\beta}^T\bm{x_i}\right).
\label{eq:sigdiscrete}
\end{equation}
Discrete-time models can be used as approximations to continuous-time ones.
This is done in \cite{zhong2019survival}, where multiple classifiers with
tied weights give the probability of patients still at risk to have an event
at each given time. In particular, when logistic regression is used,
\eqref{eq:sigdiscrete} is recovered. \cite{zhong2019survival} also
introduce a stacking method to be able to cast this model as a standard
classification model and train all its parameters at once; we employ the
same approach in our proposed method~(Sec.~\ref{methods}).

\subsection{Federated learning}
\label{back_fl}

In federated learning (FL)
\cite{shokri2015privacy,mcmahan2017communication}, one starts with data
distributed among multiple centers and strives to jointly learn a predictive
model from these data by transmitting as little information as possible from
the centers. This can be accomplished in multiple ways, depending on the
kind of model one is concerned with. For neural networks, trained \textit{via} some
variant of stochastic gradient descent (SGD), it is typically done by
introducing a centralized \emph{aggregator}, which will securely combine the
gradients produced at each center.

Specifically, given multiple datasets~$Z_k =\big\lbrace (\bm{x}_i,
y_i)\big\rbrace_{i \in I_k}$ for each center $k \in [C]$, a model $f_\theta
(\bm{x})$, and a per-sample loss~$\ell(\hat{y}, y)$ between predictions
$\hat{y}$ and actual values $y$, we can write the per-sample
\emph{separable} loss with respect to model parameters~$\theta$ as
\begin{equation}
    \mathcal{L} \left(\theta; \lbrace Z_k \rbrace_k \right) = \sum_{k = 1}^C \sum_{i \in I_k} \ell
    (f_\theta(\bm{x}_i), y_i).
    \label{eq:separable_loss}
\end{equation}

\paragraph{Basic FL algorithm}
A simple optimization algorithm to minimize this loss consists in a
distributed minibatch stochastic gradient descent. At each optimization
step $q$, each center $k$ evaluates a local gradient $g_{q, k}$ on some
local batch $B_{q, k} \subset I_k$ at the current shared value of the
parameters~$\theta_q$,
\begin{equation}
    g_{q, k} = \nabla_{\theta_q} \bigg[ \sum_{i \in B_{q, k}} \ell(f_{\theta_q}
    (\bm{x}_i), y_i) \bigg]. \label{eq:aggregate}
\end{equation}
These local gradients are then communicated to the aggregator, which
computes an aggregated gradient~$\sum_{k=1}^C g_{q, k}$. The aggregator
iterates the optimization scheme to obtain new parameters~$\theta_{q+1}$,
which are communicated back to the centers.

It is possible to choose the local batches $B_{q, k}$ to ensure that the
above steps exactly match computations performed if the datasets were
centralized. This can be done by first choosing a global batch $B_q$
uniformly from~$\cup_k I_k = [N]$, and then splitting it into multiple local
batches~$B_{q, k} = B_q \cap I_k$. Whenever we use this batch sampling
scheme, we designate the resulting basic federated learning algorithm as
\emph{pooled-equivalent FL}.

Notice that the assumption that the loss is separable, at least by center,
is key for federated learning, since communicating data points $\bm{x}_i$
between centers is restricted. Some losses, and in particular the Cox
loss~\eqref{eq:cox}, are not separable, and thus cannot be used
straightforwardly in this framework.

\paragraph{Improvements}
A number of modifications can be done to improve the above formulation of
FL. In particular, while it guarantees pooled performance is preserved, it
also raises concerns regarding the amount of communication needed, as well
as possible leakages of training data. To address the former, one might
reduce the frequency of communications while doing more local steps, leading
to a scheme called \emph{federated averaging}
\cite{mcmahan2017communication}. This comes at the expense of having no
guarantees on the convergence of the algorithm, which can be particularly
problematic when data distribution significantly changes from center to
center, \textit{i.e.} when data is heterogeneous \cite{zhao2018federated}. Another
possibility is to make gradients sparse \cite{sattler2019sparse}.

In order to prevent data leakages, techniques such as differential privacy
\cite{dwork2014algorithmic,abadi2016deep} and secure aggregation
\cite{bonawitz2017practical} can be used. From an empirical perspective,
simply adding noise or sparsifying the gradients seems to be enough to
prevent attacks seeking to recover data from gradients \cite{zhu2019deep}.

\section{Motivation}
\label{motiv}

Training a Cox PH model~\eqref{eq:coxhazard} in a federated fashion is
challenging due to the non-separability of the Cox loss. Indeed, for
estimating $\bm{\beta}$, one needs to minimize the partial
log-likelihood~\eqref{eq:cox}, where each term corresponding to
individual~$i$ involves a sum over an index set of the form~$\lbrace j | t_j
\geq t_i \rbrace$. However, at the center~$k$ where~$i$ belongs, one does
not have access to this full set, but only to its intersection with the
local dataset $I_k$.

A possibility to circumvent this problem is to approximate the risk
set~$\lbrace j | t_j \geq t_i \rbrace$ by its intersection with the local
set~$I_k$. However, this leads to a different model: if we are forced to
stratify data by center, we are not minimizing the negative log-likelihood
of the Cox PH model~\eqref{eq:cox} but of a \textit{stratified} Cox
model~\eqref{eq:coxstratlog}. While using the stratified model might be
desirable in some cases, it might also have severe consequences in terms of
performance when the distributions of the different centers are not the
same, as we show in Sec.~\ref{sec:synthetic}.

As discussed in the next section, the distributed optimization
of~\eqref{eq:cox} is possible, but has a number of drawbacks. Instead, in
Sec.~\ref{methods}, we propose working with a proxy to the Cox model, which
has a separable loss and is thus amenable to federated learning in a
plug-and-play fashion.

\section{Previous Work}
\label{previous}

In \cite{lu2015webdisco}, an algorithm named WebDISCO is introduced for
training the Cox model~\eqref{eq:coxhazard} in a distributed fashion. In
their approach, each center sends aggregate information to a central node
for a number of rounds; the central node is then able to optimize the Cox
partial log-likelihood as if all the data was pooled together. We now
briefly review their method and expose its limitations.

\paragraph{WebDISCO}
For each center $k$ and each unique event time~$s^{(m)}$, we introduce 
$D^{(m)}_k$ and $R^{(m)}_k$ as the subsets of individuals in center $k$
who had an event, and who are still at risk at time $s^{(m)}$ respectively:
\begin{align}
    D_k^{(m)} & \triangleq \left\lbrace i \in I_k \mid t_i =
        s^{(m)}, \delta_i=1 \right\rbrace, \\
    R_k^{(m)} & \triangleq \left\lbrace i \in I_k \mid t_i \geq
        s^{(m)} \right\rbrace.
\end{align}
In order for the central server to compute the gradient of the Cox partial
log-likelihood~\eqref{eq:cox}, each center $k$ communicates to the central
server the couple
\begin{equation}
\begin{cases}
\zeta(R_k^{(m)}, \bm{\beta}) = \sum_{i \in R_k^{(m)}}
\exp\left(\bm{\beta}^T\bm{x}_i\right), \\
\bm{\mu}(R_k^{(m)}, \bm{\beta}) = \sum_{i \in R_k^{(m)}}\bm{x}_i
\exp\left(\bm{\beta}^T\bm{x}_i\right),
\end{cases}
\end{equation}
for all time indices $m \in [T]$. The gradient of the Cox partial
log-likelihood \eqref{eq:cox} can then be computed as
\begin{equation}
    \nabla_{\bm{\beta}} \mathcal{L} (\bm{\beta}) = \sum_{k=1}^C \sum_{m=1}^T \sum_{i
        \in D_k^{(m)}} \bm{x}_i 
    -\sum_{m=1}^T \left(\sum_{k=1}^C |D_k^{(m)}|\right) \frac{\sum_{k=1}^C
       \bm{\mu}(R_k^{(m)}, \bm{\beta})} {\sum_{k=1}^C \zeta(R_k^{(m)}, \bm{\beta})}.
\end{equation}
Analogous formulas are derived to allow the central server to
build the Hessian thanks to quantities communicated by each center.

\paragraph{Limitations}
Compared to our work, WebDISCO is limited in the following aspects. In
terms of bandwidth requirements, each center has to send, at each
optimization step, $\mathcal{O}(TP)$ terms for the gradients, and an
additional~$\mathcal{O}(TP^2)$ for the Hessian if a second-order
gradient descent is carried
out, whereas our method only requires the exchange of~$\mathcal{O}(T+P)$ terms at
each communication step. In terms of generality, the scope of the
algorithm is limited to the linear Cox model; it cannot be adapted, for
instance, to neural networks, which is the case for our method. In
terms of privacy, it might leak sensitive information, as we detail in
the next paragraph.

\paragraph{Privacy leakage}
In WebDISCO, the aggregate information that is sent by each center can
reveal individual-level data. Notice that, by telescoping the
communicated quantities with respect to~$m$, the central server can
infer $\zeta(D_k^{(m)}, \bm{\beta})$ and $\bm{\mu}(D_k^{(m)},
\bm{\beta})$, which are aggregated measurements over a potentially much
smaller set $D_k^{(m)}$, consisting only of patients who had an event at
the given time.

If one such set $D_k^{(m)}$ contains only one event, then the measurements
of the related individual~$\bm{x}_i$ are exposed to the server. Indeed,
in this case, the fraction $\bm{\mu}(D_k^{(m)}, \bm{\beta}) /
\zeta(D_k^{(m)}, \bm{\beta})$ is equal to~$\bm{x}_i$.

If no such set contains a single element, the measurements of the related
individuals are still at risk. Indeed, because an iterative
optimization is carried out, the central server observes~$\left(
\zeta(D_k^{(m)}, \bm{\beta}), \bm{\mu}(D_k^{(m)}, \bm{\beta}) \right)$
for multiple measurements of $\bm{\beta}$. This can be seen as a system
of nonlinear equations where the unknowns are the $\left \lbrace
\bm{x}_i\right \rbrace_{i \in D_k^{(m)}}$, and for which numerical
solutions can be attempted.

\paragraph{Related survival models}
Survival analysis is not confined to CoxPH models:
other popular survival models include
accelerated failure time (AFT)
methods~\cite{wei1992accelerated, ranganath2016deep}
or non-parametric sampling-based
methods~\cite{chapfuwa2018adversarial, miscouridou2018deep}.
In the following, we focus on the CoxPH model as it is one
of the most popular survival models,
especially in the medical
field~\cite{wang2019machine, wang2020remdesivir}.

\section{Methods}
\label{methods}

\subsection{Discrete-time extended stacked Cox model}
\label{model}

\paragraph{Discrete-time setting} 
In the following, we consider the discrete-time approach introduced in
Sec.~\ref{back_surv}. All observed times~$\lbrace t_i \rbrace_{i \in
[N]}$ are binned to a finite set $\lbrace s^{(m)} \rbrace_{m \in [T]}$ of
size $T$. Note that this finite set may contain less values than the number
of unique event times if the latter are first quantized. Such a
quantization is performed by first choosing a binning interval $Q$ and
applying the mapping~$t \mapsto Q \ceil{t / Q}$ to observed times~$t_i$.

\paragraph{Extended Cox model}
Given parameters $\bm{\alpha} = (\alpha^{(1)}, \dots, \alpha^{(T)})$,
$\bm{\beta} \in \mathbb{R}^{P'}$ and a mapping~$\bm{\phi}_\theta:
\mathbb{R}^P \to \mathbb{R}^{P'}$ parametrized by $\theta$, we model the
conditional probability of an event occurring at time~$s^{(m)}$ for
individual $i$ by
\begin{equation}
\label{eq:model}
\begin{array}{rcl}
    p^{(m)}_{\bm{\alpha}, \bm{\beta}, \theta}(\bm{x}_i) &=& \mathbb{P} \left[
        \tau_i = s^{(m)} | \tau_i \geq s^{(m)} \right] \\
    &=& \sigma \left( \alpha^{(m)} + \bm{\beta}^T \bm{\phi}_\theta
    (\bm{x}_i) \right).
\end{array} 
\end{equation}
This model is a \emph{discrete-time proportional hazards} model which
extends the Cox model~\eqref{eq:coxdt} with an intermediate representation
$\bm{\phi}_\theta (\bm{x}_i)$. When $\bm{\phi}_\theta (\bm{x}_i) =
\bm{x}_i$, one retrieves the discrete-time Cox model. In general, however,
one can employ more complex models such as neural networks.

\paragraph{Stacking method}
A simple method recasting~\eqref{eq:model} as a classification problem can
be used to learn all the time-varying biases $\bm{\alpha}$ at once
\cite{zhong2019survival}. For each sample $(\bm{x_i}, t_i)$, we first
introduce the binary labels
\begin{equation}
    y_i^{(m)} = 
    \begin{cases}
    0 & \text{if } t_i > s^{(m)} \\
    \delta_i & \text{if } t_i = s^{(m)},
    \end{cases}
\end{equation}
for all $m$ such that $s^{(m)} < t_i$ and additionally, if $\delta_i = 1$,
for $m$ such that $t_i = s^{(m)}$. Notice that by
construction~$\mathbb{P}[y_i^{(m)} = 1 ] = \mathbb{P}[\tau_i = s^{(m)} |
\tau_i \geq s^{(m)}]$.

For each original sample, we generate new samples
\begin{equation}
    \bm{x}_i^{(m)} = \left[ \bm{e}^{(m)}, \bm{x}_i \right],
\end{equation}
where $\bm{e}^{(m)} \in \mathbb{R}^T$ is the $m$-th vector of the canonical
basis in dimension $T$. Each of the samples~$\bm{x}_i^{(m)}$ is associated
to label $y_i^{(m)}$.

\begin{figure}
    \centering
    \includegraphics[width=0.45\textwidth]{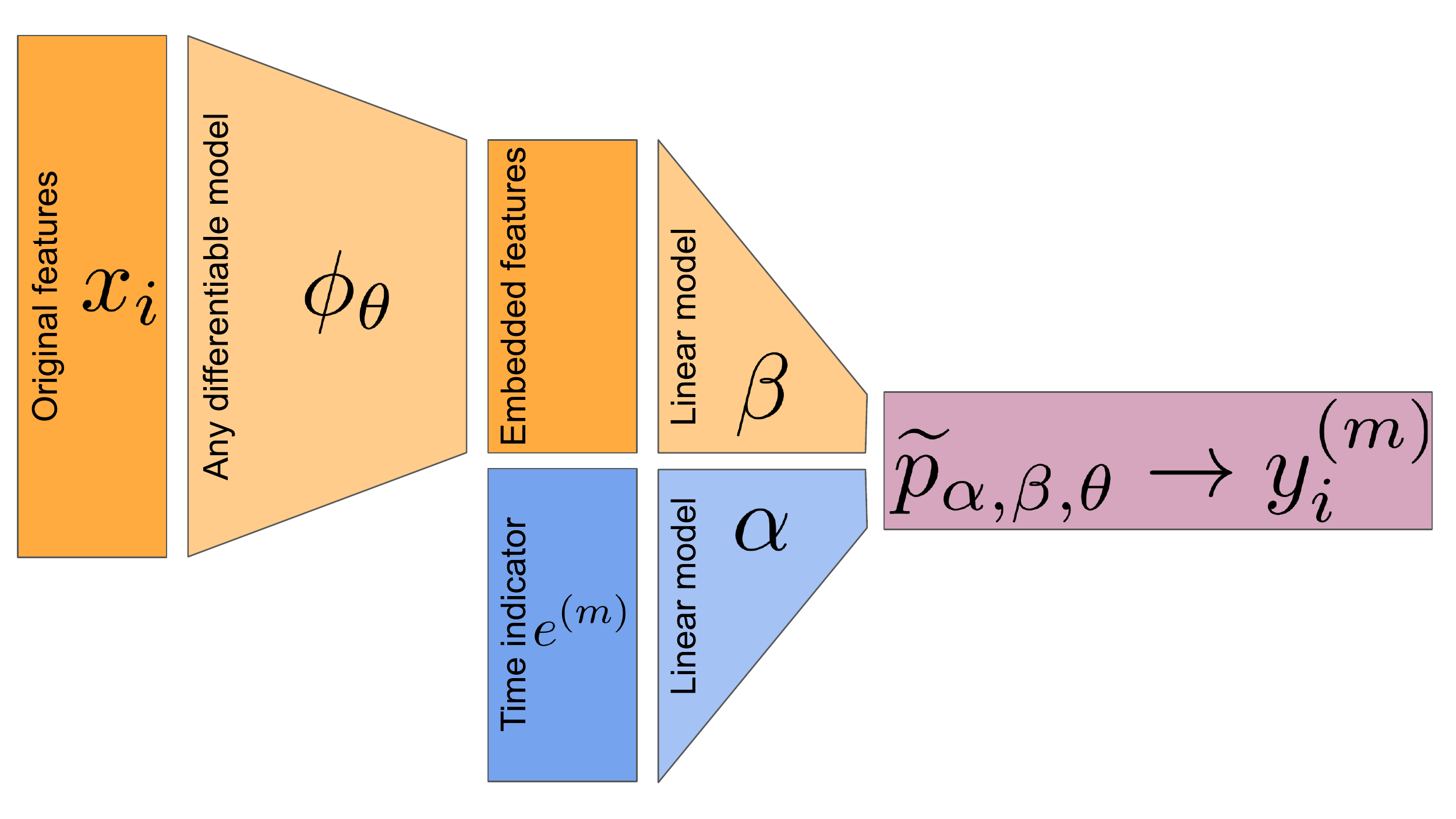}
    \caption{Parametric model $\widetilde{p}_{\bm{\alpha}, \bm{\beta}, \theta}(x_i^{(m)})$
    to represent survival probabilities using the stacking method
    described by~\eqref{eq:model_trick}.}
    \label{fig:network}
\end{figure}

The classification task associated to the \textit{stacked} dataset $\lbrace
(x_i^{(m)}, y_i^{(m)}) \rbrace_{i, m}$ corresponds to the occurrence of an
event at the time encoded in the canonical vector, provided no event
occurred at previous times. As a consequence, if we define a binary
classification model on the stacked dataset as
\begin{equation}
\label{eq:model_trick}
\begin{array}{rcl}
    \widetilde{p}_{\bm{\alpha}, \bm{\beta}, \theta}(\bm{x}_i^{(m)})
    &=&
    \sigma\left(
    [\bm{\alpha}; \bm{\beta}]^T [\bm{e}^{(m)}; \bm{\phi}_{\theta}(\bm{x}_i)]
    \right) \\
    &=& \mathbb{P}[y_i^{(m)} = 1],
\end{array}
\end{equation}
where $[\cdot; \cdot]$ denotes vector concatenation, then this model
satisfies
\begin{equation}
    p^{(m)}_{\bm{\alpha}, \bm{\beta}, \theta}(\bm{x}_i) =
    \widetilde{p}_{\bm{\alpha}, \bm{\beta}, \theta}(\bm{x}_i^{(m)}).
\end{equation}
In other words, fitting~\eqref{eq:model_trick} is equivalent to
fitting~\eqref{eq:model}. Model \eqref{eq:model_trick} is sketched in Fig.
\ref{fig:network}.

\paragraph{Loss function}
Thanks to the stacking method, the model parameters $(\bm{\alpha},
\bm{\beta},\theta )$ can be found by training model~\eqref{eq:model_trick}
on the stacked dataset with a standard classification loss. More precisely,
we minimize the binary cross-entropy loss between the model
$\widetilde{p}_{\bm{\alpha}, \bm{\beta}, \theta}(\bm{x}_i^{(m)})$ and the
true labels $y_{i}^{(m)}$,
\begin{align}
    \mathcal{L}_\text{BCE} (\{\bm{\alpha}, \bm{\beta}, \theta\};
        \{\bm{x}_{i}^{(m)}, y_{i}^{(m)}\}_{i, m}) =
    \sum_{i = 1}^n \sum_{
        \substack{
        m \\ ~t_i \geq s^{(m)}
        }
    }
    \ell_\text{BCE} \big( \widetilde{p}_{\bm{\alpha}, \bm{\beta},
        \theta}(\bm{x}_i^{(m)}), y_{i}^{(m)} \big),
    \label{eq:our_loss}
\end{align}
where $\ell_\text{BCE} (\hat{y}, y) = -y \log \hat{y} - (1 - y) \log \big(1 -
\hat{y}\big)$. This loss function can be easily and efficiently implemented
in generic deep learning frameworks, opening the possibility of using
advanced deep learning architectures for the intermediate representation
$\bm{\phi}_{\theta}$.

\subsection{Federated learning}

We now discuss the properties of the proposed model in the federated
setting.

\paragraph{Federated optimization}
An important property of our loss~\eqref{eq:our_loss} is that it is
separable with respect to individual \textit{stacked} samples. As a
consequence, once the local stacked datasets have been built, it is
straightforward to optimize our proposed model using the federated learning
approach described in Sec.~\ref{back_fl}.

\paragraph{Stacking in the federated setting}
Prior to the federated optimization, each center needs to build the stacked
datasets. Each center can do it individually provided the unique times~$s^{(m)}$
are known. As a consequence, all centers must first agree on the
set $\lbrace s^{(m)} \rbrace_{m \in [T]}$ of unique times.

Such an agreement can be reached easily since we use quantized times with a
sampling period~$Q$, as discussed in Sec.~\ref{model}. Given $Q$, it is
sufficient to compute the maximum event time among all centers for all
individual centers to define a common grid. A maximum event time can be
found by computing the maximum of all the local maxima. This computation
can be performed either through the server or \textit{via} multi-party
computation~\cite{kolesnikov2009maxmpc}, thereby ensuring privacy of the
local event times.

\paragraph{Computation and communication costs}
At each communication round of a given federated learning algorithm, all the
parameters need to be exchanged, which scales
as~$\mathcal{O}(T + P' + |\theta|)$.
When no intermediate representation is used, this is equal
to~$\mathcal{O}(T + P)$,
which is much lower than the~$\mathcal{O}(TP)$ required in the
previously proposed WebDISCO algorithm.

In terms of computation costs, the stacking approach increases the number of
samples, which is now upper-bounded by $NT$ instead of the original $N$.
Therefore, one has interest in choosing the sampling period $Q$ such that
$T$ is small to prevent detrimental effects on the total computation time.

\paragraph{Privacy}
Although the stacked datasets can be built in a
privacy-preserving fashion, the updates sent by the centers can
leak information about their local datasets, as well as the central model.
This is where the separability of the proposed model provides an important
advantage: generic privacy-preserving methods for federated learning can be
readily used for our model. For WebDISCO, on the other hand, further
reflection is needed before applying generic privacy-preserving solutions
such as differential privacy.

\subsection{Learning schemes}

\label{sec:learning_schemes}
In our experiments, we compare the performance of different learning schemes
on multiple datasets. A scheme consists in both a model and a procedure for
training it.

\paragraph{Baselines}
Our baselines consist in training the Cox model~\eqref{eq:coxhazard} in a
non-distributed fashion. In particular, we evaluate i) the \emph{pooled}
performance (\textbf{POOL}), that of the model trained with all data pooled
together; ii) the average \emph{local} performance (\textbf{LOCAL}), that of
the models trained at each center separately; and iii) the \emph{ensembled}
performance (\textbf{ENS}), that obtained by averaging the predictions of
the models trained at each center. These models are trained using the
\texttt{lifelines} library~\cite{cameron_davidson_pilon_2020_lifelines}
implementation of the Newton method.
We stress that, in terms of c-index, the \textbf{POOL} method is equivalent
to the baseline WebDisco method~\cite{lu2015webdisco}.

Another baseline, which is used in Sec.~\ref{sec:realworld}, comes from
training on pooled data, but using a \emph{boosting} approach instead,
specifically XGBoost \cite{chen2016xgboost} with the Cox partial
log-likelihood as the objective function. In our experience, boosting
typically gives similar or better results than linear methods in this
context, and as such we use it as baseline when working with the TCGA
datasets.

\paragraph{Cox in minibatches (MINI)}
This method consists in the optimization of the Cox loss~\eqref{eq:cox} by
means of minibatch SGD, or more generally a variant such as Adam. Notice
that the risk set for each individual is built by only considering samples
present in the batch, but that each batch is sampled from the pooled data.

\paragraph{Stratified/na\"{i}vely-federated Cox (N-FL)} 
Optimization of the Cox loss stratified by center~\eqref{eq:coxstratlog}.
This is done following the procedure in Sec.~\ref{back_fl}, \textit{i.e.} by
combining gradients coming from the different centers. Risk sets are thus
built considering only samples present in the batch \emph{and} coming from
the same center as the individual. We denote this scheme \emph{na\"{i}ve FL}
in our experiments.

\paragraph{Model~\eqref{eq:model_trick} with pooled-equivalent FL (DT-FL)}
We simulate this optimization of the loss~\eqref{eq:our_loss} being
done on separate centers with minibatches. However, since the loss
is separable, one should obtain the same results if data is pooled
together, see discussion in Sec.~\ref{back_fl}.
For~$\bm{\phi}_\theta$, we choose either the identity function, in
which case the discrete-time Cox model is recovered
(Sec.~\ref{sec:synthetic} and~\ref{sec:realworld}), or a more complex
neural network adapted to the task in hand (Sec.~\ref{sec:histo}).
We denote this scheme \emph{discrete-time FL} in our experiments.

\subsection{Performance evaluation}

We evalute the performance of each scheme by using cross-validation (CV) in
two different ways. In the first, each center splits their data in 5 folds,
and at the $i$-th round of the CV, all the $i$-th folds are combined
together in order to build the test set. This procedure provides a proxy to
how the model generalizes at the different centers, on average. As this
computation might not be feasible in practice, we also study the
\emph{out-of-center} (OOC) generalization, by setting aside, at each round
of CV, all data points of a given center.

The metric used to evaluate the models is the concordance index
\begin{equation}
    \text{c-index} = \underset{\substack{i:\delta_i=1 \\ j:t_j >
    t_i}}{\mathbb{E}} \mathbbm{1}_{[\eta_j < \eta_i]},
\end{equation}
where $\eta_i$ is a risk score assigned to each individual---a high
concordance index implies that risk scores and times-to-event are inversely
ranked. For proportional hazards models, one can use as score the
time-independent part of the hazard function. Thus, for the Cox
models~\eqref{eq:coxhazard} and~\eqref{eq:sigdiscrete},~$\eta_i =
\bm{\beta}^T \bm{x}_i$, and, for model~\eqref{eq:model_trick}, $\eta_i =
\bm{\beta}^T \bm{\phi}_\theta (\bm{x}_i)$.

\begin{figure*}[ht]
    \centering
    \includegraphics[width=0.48\textwidth]{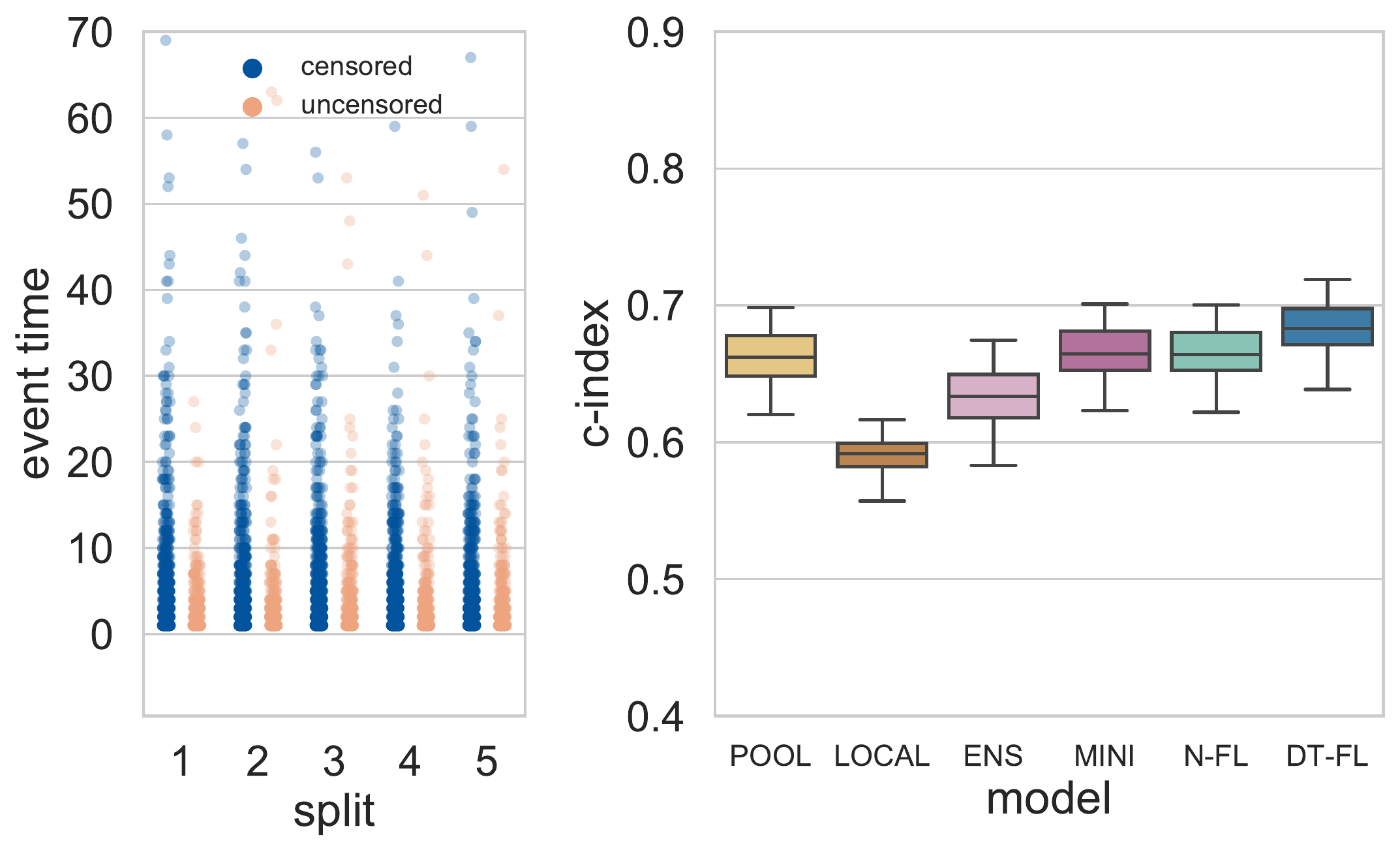} \;
    \includegraphics[width=0.48\textwidth]{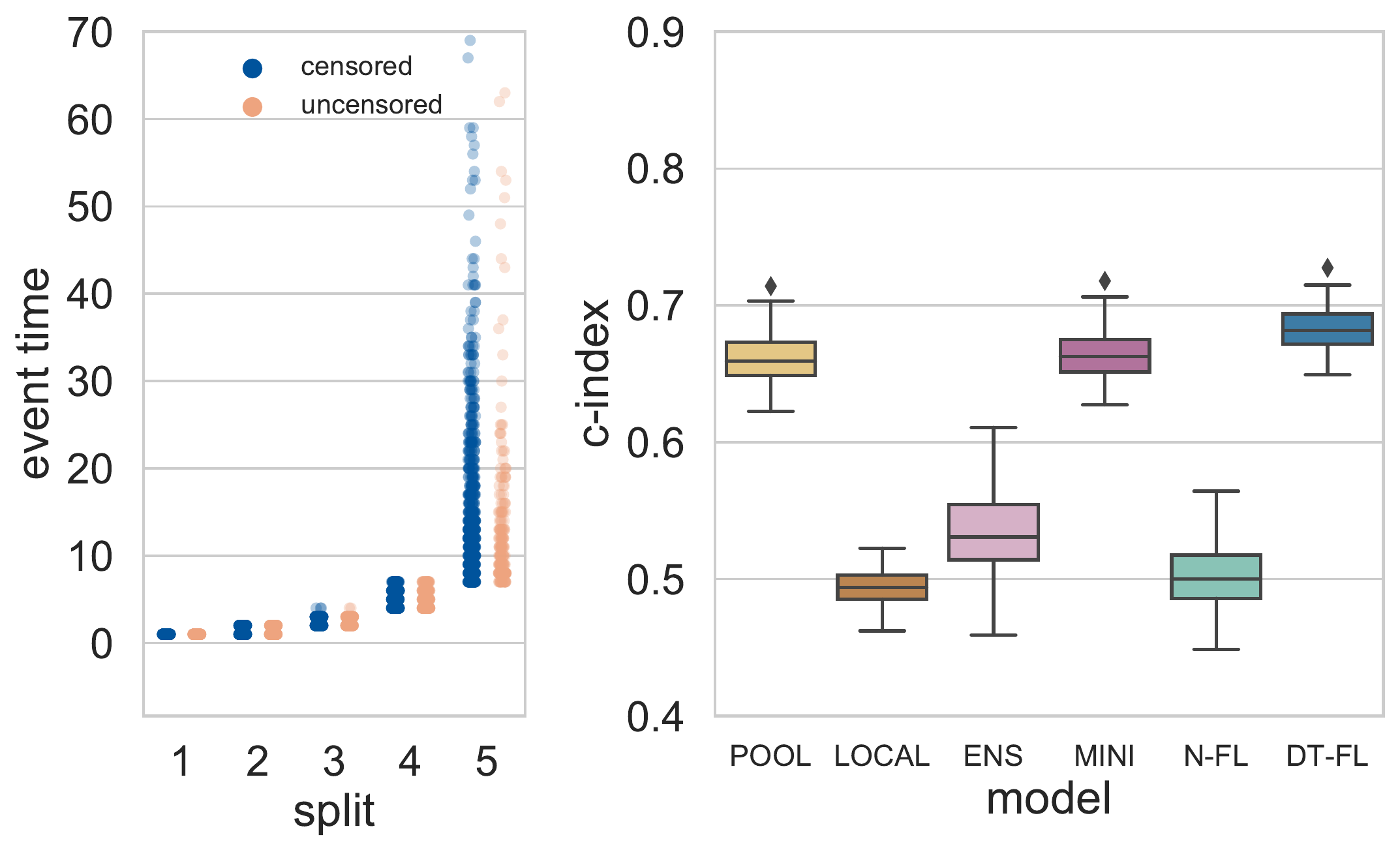}
    \caption{Experiments on synthetic data for learning schemes described in
    Sec.~\ref{sec:learning_schemes}, generated according to the procedure
    described in Sec.~\ref{sec:synthetic}. \emph{Left:} performance of
    different schemes given synthetic data which is uniformly distributed
    across centers. Distribution of splits can be seen on the left.
    \emph{Right:}
    analogous to the left figure, but given synthetic data ordered according
    to endpoint. Na\"{i}ve federated scheme performance decreases
    considerably in this case, whereas discrete-time federated is not
    significantly affected.}
    \label{fig:synthetic}
    \vskip-2ex
\end{figure*}

\section{Results and Discussion}
\label{results}

\subsection{Experiments on synthetic tabular data}
\label{sec:synthetic}
We start by showing the effects of the different schemes on
synthetically-generated data.  The features for each individual are
generated by sampling from a normal distribution with
variance~$1/P$,~$\bm{x}^{(i)} \sim \mathcal{N} (0, P^{-1} I_P)$,
thereby ensuring that~$\mathbb{E}\| \bm{x}^i \|^2 = 1$.
Observed times $t_i$ are then set
according to the following procedure.  We first sample \emph{actual times}
$\tau_i$ from a Cox model with constant baseline hazard, so that its c.d.f.
is given by $\Psi(u) \triangleq \mathbb{P} [\tau < u] = 1 - e^{-u
\exp(\beta^T \bm{x}_i)}$; \emph{censoring times} are then sampled from an
uniform distribution $U\big(0, \Psi^{-1} (\frac12)\big)$.  Finally, the
observed times are defined as the minimum of censoring and actual times.

For this example, we simulate data distributed across $C = 5$ centers with
$n = 1000$ patients each, where each patient is described by $P = 200$
features. Experiments are performed either with data being uniformly
distributed across centers, or stratified according to endpoint, see
Fig.~\ref{fig:synthetic}.  We only perform 5-fold cross validation, as in
the uniform case average and OOC performances are the same, and in the
stratified case there is almost no OOC generalization. Displayed results are
averages over 20 runs of 5-fold CV.
Experimental details are provided in App.~\ref{app:experimental_details_global}.

We split the data across centers in two different ways. For the first
experiment, we generate $N = Cn$ samples according to the above procedure
and then split them at random, assigning $n$ of them to each center. As
seen in Fig.~\ref{fig:synthetic}, left, all methods perform similarly, apart
from local and ensemble which, as expected, give subpar results.
Note that as the underlying Cox models for the pooled
and discrete-time FL methods differ,
we do not expect their results to be exactly equal.

As a second experiment (Fig.~\ref{fig:synthetic}, right), we order the
samples per time of event $t_i$ before splitting them, so that the
smallest~$t_i$ go to the first center and the largest to the last. Doing
this effectively destroys the performance of the na\"{i}ve FL procedure,
which minimizes the stratified loss~\eqref{eq:coxstratlog}. Most
importantly, the performance of discrete-time FL remains unaltered---indeed,
\emph{it is not dependent on how the data is split across centers}. This is
one of the main advantages of our approach.
\begin{figure*}[ht]
    \centering
    \includegraphics[width=0.42\textwidth]{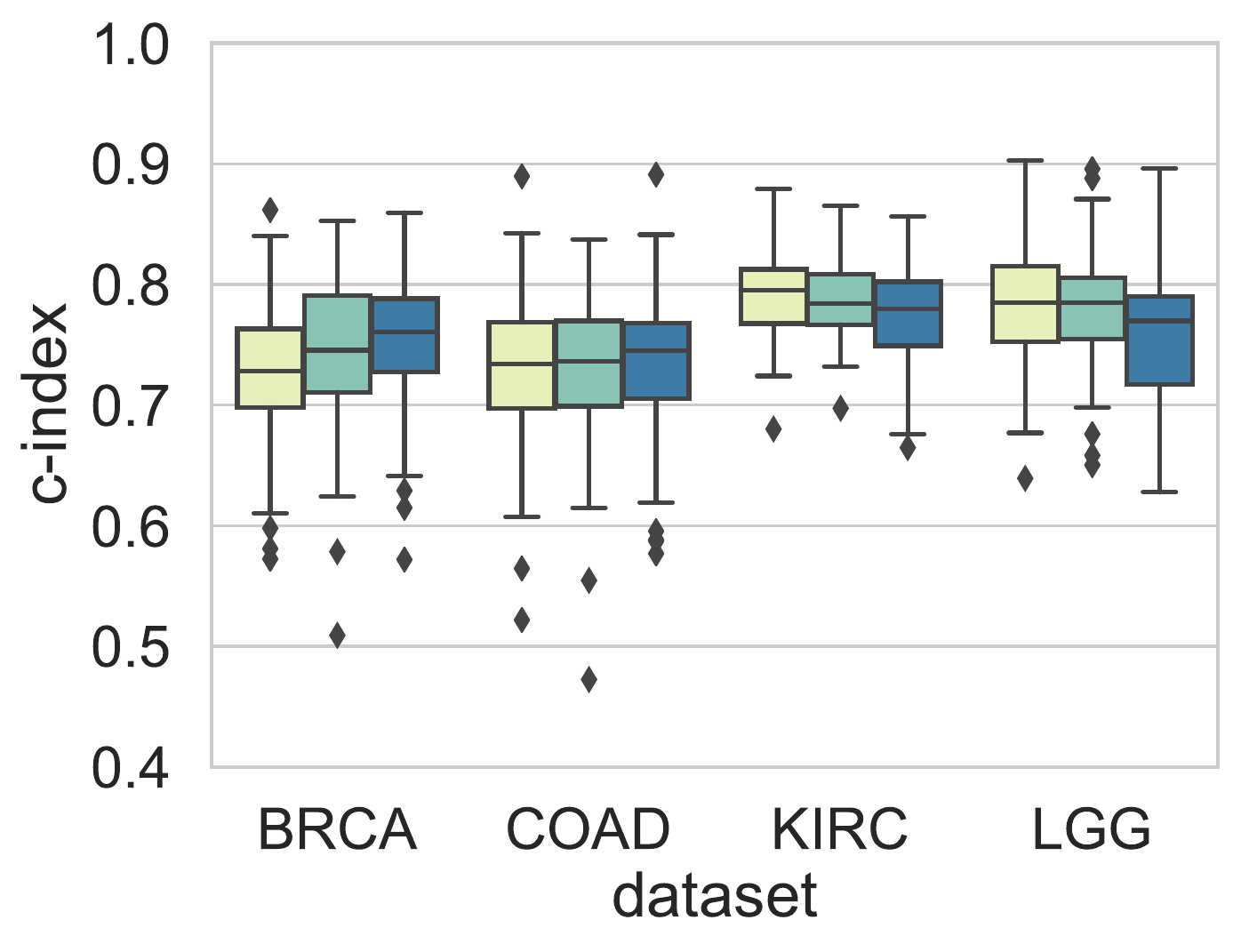} \kern2em
    \includegraphics[width=0.42\textwidth]{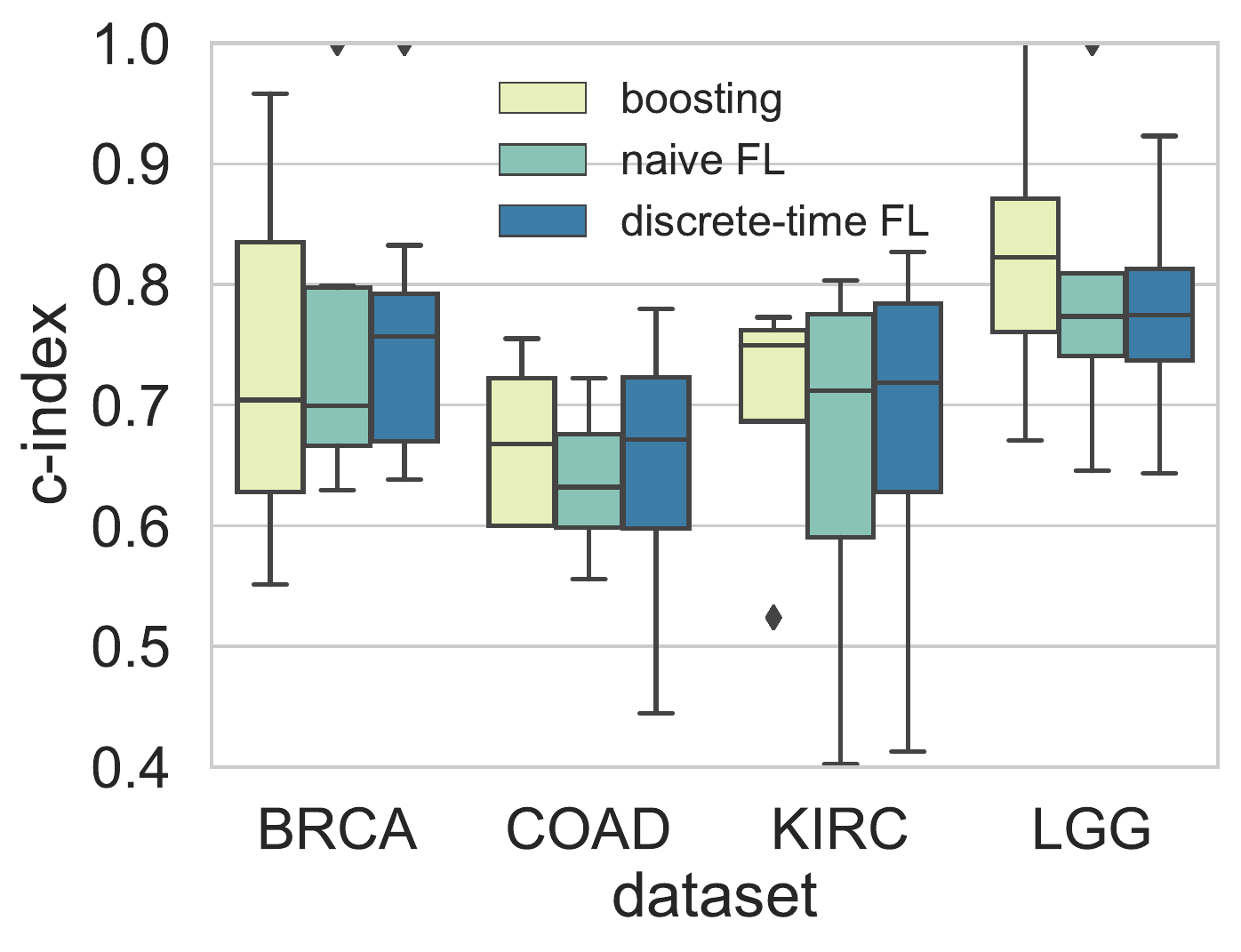}
    \caption{Experiments on 4 different TCGA tabular datasets, each spread
    across 4 to 6 centers depending on the dataset.
    Different approaches are compared,
    either using 5-fold cross validation with the test fold coming
    from the different centers
    (left), or in a out-of-center fashion, with the whole center data used
    for testing (right). Experiments do not reveal any significant
    difference in the performance of the different approaches, and
    discrete-time FL provides a reasonable performance in all cases.}
    \label{fig:tcga}
    \vskip-2ex
\end{figure*}
\subsection{Experiments on TCGA tabular datasets}
\label{sec:realworld}
For the experiments on real-world data, we use datasets from The Cancer
Genome Atlas (TCGA). For each type of cancer with more than 200 patients, we
download clinical information and preprocess datasets separately,
transforming the categorical variables into dummy variables and removing
columns with more than $95\%$ of missing values. We use the curated TCGA
Pan-Cancer Clinical Data Resource (TCGA-CDR) \cite{liu2018integrated} to
obtain the overall survival time and event status for each patient.

We first train baseline boosting models described
in Sec.~\ref{sec:learning_schemes} on the generated features and
labels, and keep the four cancer types with the highest average score
obtained in 50 cross-validation runs. The four selected cancer
types are: Breast Invasive Carcinoma (BRCA), Colon Adenocarcinoma (COAD),
Kidney Renal Clear Cell Carcinoma (KIRC), and Low Grade Glioma (LGG).
Details can be found in Tab.~\ref{tab:datasets}.
\begin{table}[ht]
    \centering
    \caption{TCGA tabular datasets used in the experiments.} \vskip1ex
    \begin{tabular}{p{15mm}p{18mm}p{18mm}p{15mm}}
    \toprule
    Cancer\newline type & Number of\newline patients $N$ & Number of\newline
            covariates $P$ & Censoring\newline percentage \\ \midrule
    BRCA & 1096 & 41 & 86.2\% \\
    COAD & 458 & 47 & 77.7\% \\
    KIRC & 537 & 29 & 67.0\% \\
    LGG  & 514 & 20 & 75.7\% \\ \bottomrule
    \end{tabular}
    \label{tab:datasets}
\end{table}

The TCGA patient's barcode is used to identify unique patients and to map
data collection to a specific tissue sort site (TSS). We identify 64
different sites, many of which contain very few samples.  We thus regroup
these sites in at most 6 different centers, corresponding to geographic regions: USA
(Northeast, South, Middlewest, West), Canada and Europe---countries
from outside these regions do not have a significant amount of samples.
Per-center details are provided in App.~\ref{app:details_tcga_datasets}.

For these 4 datasets, each split by the aforementioned centers, we evaluate the
performance of 3 different schemes: boosting on the pooled data, na\"{i}ve
FL for minimizing the stratified Cox loss~\eqref{eq:coxstratlog}, and
discrete-time FL~\eqref{eq:our_loss} with~$\bm{\phi}_\theta$ equal to
identity.  Cross-validation is performed in two different ways: either by
5-fold CV, with test folds coming from all centers (Fig.~\ref{fig:tcga},
left); or in an out-of-center fashion, with the whole center data used for
testing (Fig.~\ref{fig:tcga}, right).  Experiments do not reveal significant
differences between approaches, and discrete-time FL provides reasonable
performances in all cases.
\subsection{Experiments on TCGA histopathology datasets}
\label{sec:histo}
Finally, we demonstrate how this approach can be used also with more
complex data and models by training a deep learning model on the
digitally scanned
histopathology slides that accompany the TCGA tabular data.
We use the same datasets as in the
previous subsection, up to small variations due to
the absence of slides for some patients.

Histopathology slide images are very large in size, with up to $10^5 \times 10^5$ pixels.
We preprocess them with a standard pipeline
\cite{durand2016weldon,courtiol2018classification},
first extracting matter-bearing, non-overlapping tiles
from these images, then passing these tiles through an ImageNet-pretrained ResNet-50,
and finally linearly autoencoding the penultimate ResNet layer's output in order to decrease the number of features.
Each slide thus results in an array of size
$8000 \times 256$, where~$8000$ is the number of tiles extracted from the
slide, and $256$ the number of features per tile.

For the function $\bm{\phi}_\theta (\bm{x}_i)$, we use a sequence of two 1D
$1 \times 1$ convolutions separated by a leaky ReLU, with the output of the second
convolution being averaged over all dimensions. Similar architectures have
been used in order to analyze histopathology data
\cite{durand2016weldon,courtiol2018classification}. The baseline is provided
by training this exact same model on pooled data, but with a Cox loss instead, i.e. the
\emph{minibatch} method previously introduced.
For clarity and due to the size of the dataset,
we do not consider the pooled, local and ensemble baselines.
Experimental details on datasets, preprocessing, and networks used
can be found in App.~\ref{app:experimental_details_global}.
\begin{table}[ht]
    \centering
    \caption{C-index obtained on TCGA histopathology slides by both the
    discrete-time FL approach and the pooled baseline. Shown are mean and
    standard deviation over 5 runs of 5-fold CV.} \vskip1ex
    \begin{tabular}{p{20mm}p{22mm}p{25mm}} \toprule
    Cancer type & MINI & DT-FL \\ \midrule
    BRCA & $0.55 \pm 0.06$ & $0.57 \pm 0.06$ \\ 
    COAD & $0.62 \pm 0.07$ & $0.65 \pm 0.07$ \\
    KIRC & $0.66 \pm 0.05$ & $0.69 \pm 0.04$ \\
    LGG  & $0.67 \pm 0.05$ & $0.68 \pm 0.06$ \\ \bottomrule
    \end{tabular}
    \label{tab:results}
\end{table}

Tab. \ref{tab:results} shows that, when applied to training 
state-of-the-art deep neural networks, discrete-time FL yields similar
performance to minibatch optimization on pooled data.
We note that for the COAD and KIRC cancer types discrete-time FL even slightly outperforms the baseline.
Yet, this difference is not always significant, notably for the BRCA and LGG cancer types.
We suspect that this slight improvement results from the difference between
underlying models, which are only asymptotically equivalent, and, in particular, from the regularization introduced by the 
time-to-event quantization.

\section{Conclusion}
\label{conclusion}

This paper investigates training survival models in a federated setting.
Our results show that na\"{i}vely federating a Cox PH model amounts to
training a stratified Cox PH model, which can have adverse effects of
performance if done unchecked.  Our proposed approach builds upon a
discrete-time Cox model trained by stacking, which gives the same results in
both federated and pooled settings.  This approach compares favorably to
previous works in terms of communication efficiency, generality and privacy.

This paper opens many future research directions.  Since the proposed model
can be directly cast in federated learning frameworks, it  is natural to
study the impact of privacy-preserving techniques, such as differential
privacy, on the performance of the resulting models.  Similarly, one could
study the impact of gradient compression techniques to further improve its
communication efficiency.

\section*{Acknowledgments}
We would like to thank Félix Balazard for pointing out the connection
between the na\"{i}ve FL approach and the stratified Cox model. We also
appreciate all the relevant feedback and comments from Owkin's Lab, in
particular from Michael Blum, Pierre Courtiol, Eric Tramel and Mikhail
Zaslavskiy.

The results shown in this paper are in whole or part based upon data generated by the TCGA Research Network: \url{https://www.cancer.gov/tcga}.

\bibliography{refs}
\bibliographystyle{plain}

\appendix
\newpage
\section{Link between continuous-time and discrete-time models}
We now explain how discrete-time models
like~\appendixcontext{\eqref{eq:model_trick}}{(18)} approximate
the continuous-time Cox model~\appendixcontext{\eqref{eq:coxhazard}}{(2)},
following the arguments of~\cite{zhong2019survival}.

The contribution to the loss of an event happening to
individual~$i$ at time $t_i = s^{(m)}$ is
\begin{equation}\label{eq:dt_contrib}\appendixcontext{}{\tag{22}}
    \alpha^{(m)} + \bm{\beta}^T \phi_\theta (\bm{x}_i) -
        \sum_{j : t_j \geq t_i} \log (1 + e^{\alpha^{(m)} + \bm{\beta}^T
        \phi_\theta (\bm{x}_j)}).
\end{equation}
Optimizing over $\alpha^{(m)}$ gives
\begin{equation}\appendixcontext{}{\tag{23}}
    \alpha^{(m)}_* = -\log \sum_j \frac{e^{\bm{\beta}^T \phi_\theta(\bm{x}_j)}}{1 +
        e^{\alpha_*^{(m)} + \bm{\beta}^T \phi_\theta (\bm{x}_j)}}.
\end{equation}
If one assumes $\alpha^{(m)}_* +
\bm{\beta}^T \phi_\theta(\bm{x}_j)$ to be very negative, then $1 + e^{\alpha^{(m)}_*
+ \bm{\beta}^T \phi_\theta(\bm{x}_j)} \approx 1$, and $\alpha^{(m)}_* \approx
-\log \sum_j e^{\bm{\beta}^T \phi_\theta (\bm{x}_j)}$. Replacing this
expression in~\eqref{eq:dt_contrib} while setting~$\phi_{\theta}$ to the identity function
shows the contribution of each event is effectively the same as that of the
Cox loss~\appendixcontext{\eqref{eq:cox}}{(3)}.

\section{Datasets from The Cancer Genome Atlas}
\label{app:details_tcga_datasets}
We provide additional details on the datasets extracted from The Cancer Genome Atlas (TCGA).
We distinguish between the tabular and image data as both datasets do not overlap entirely, and yield
slightly different counts.
These variations do not exceed $5\%$ in the total number of patients in any cancer dataset.

Centers correspond to geographic regions obtained by looking up at TCGA metadata~\cite{liu2018integrated}.
We restrict ourselves to 6 geographic regions, 4 of them belonging to the USA (Northeast, South, West, and Middlewest), 1 for Europe, and 1 for Canada, and ignore the remaining ones as they do
not bring enough patients.

For both image and tabular data, we use the same event times for each relevant patient, and restrict ourselves to centers for which data points are available.
Further, we remove centers for which only a single entry is available, as it prevents one from carrying out any meaningful intra-center cross-validation.
As a consequence, depending on the cancer dataset, the number of centers varies between $3$ and $6$.

\begin{figure}[h!]
\centering
\includegraphics[scale=0.2]{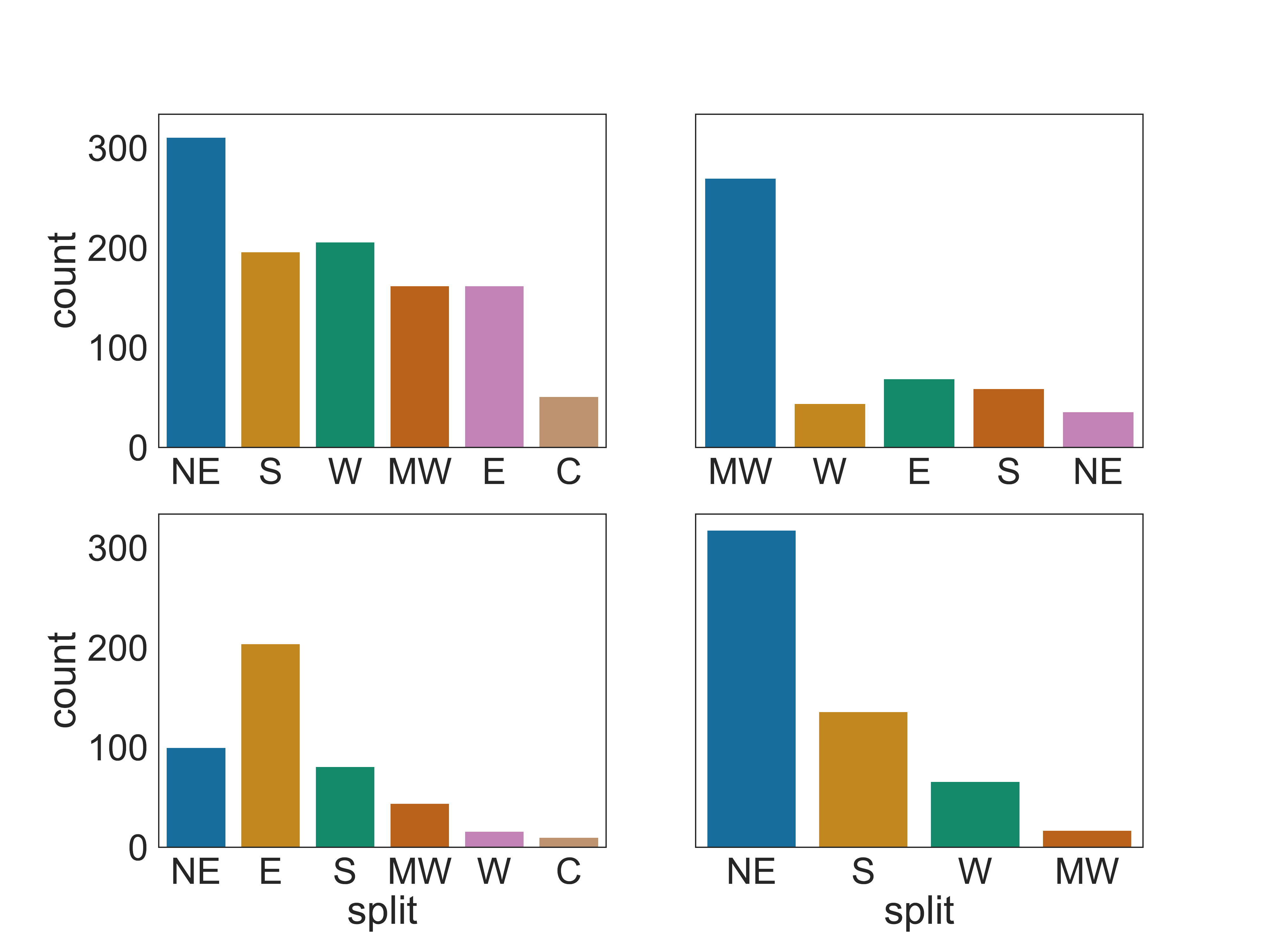}
\includegraphics[scale=0.2]{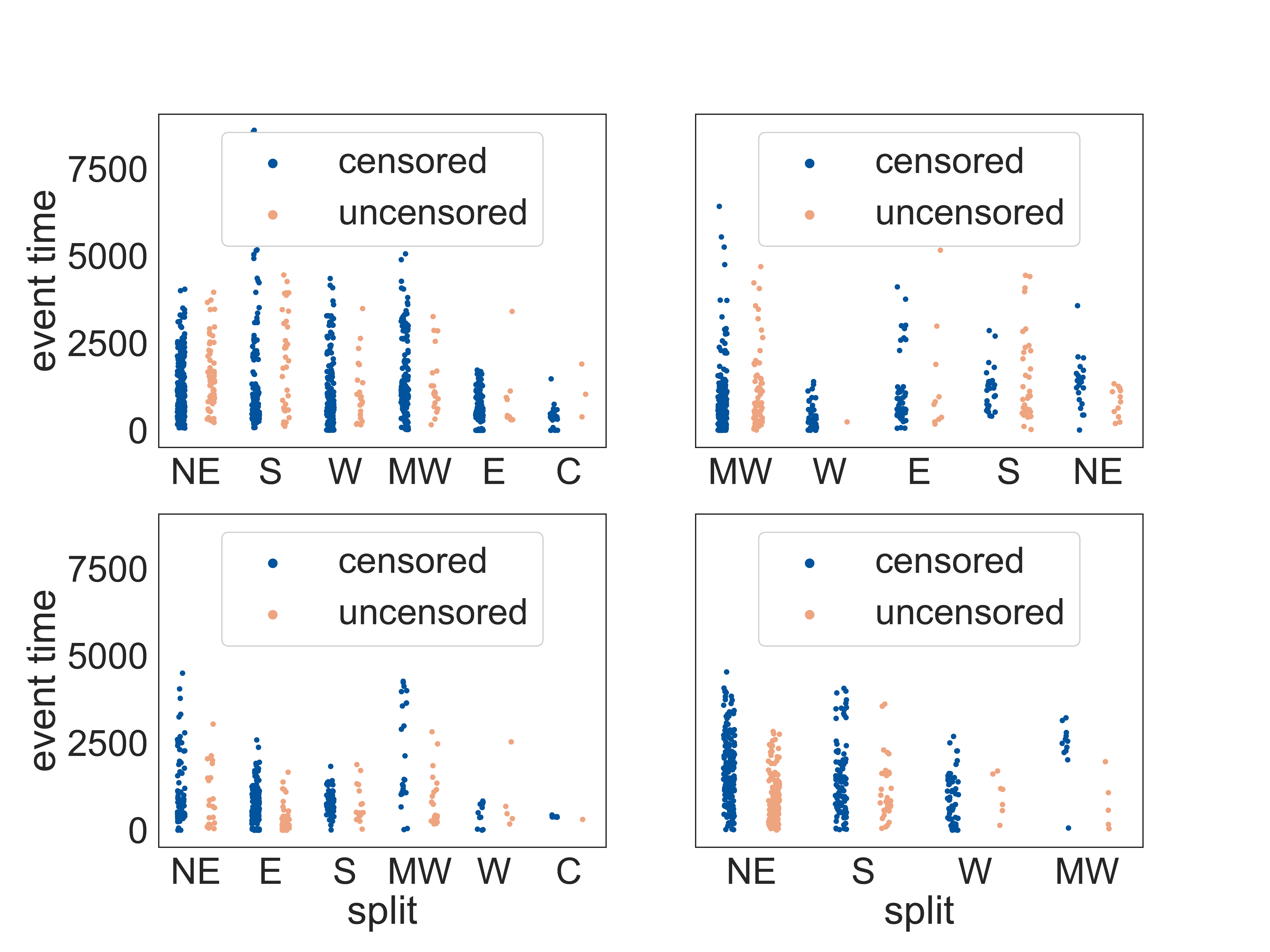}
\caption{Details on the tabular datasets from TCGA. In each figure, from left to right and top to bottom, the cancer types are BRCA, LGG, COAD and KIRC. \textit{Left:} Number of patients per center. \textit{Right:} Distribution of event times per center. NE, S, W, MW, E, C respectively stand for Northeast, South, West, Middlewest, Europe and Canada.}
\label{fig:tcga_details_tabular}
\end{figure}

\begin{figure}[h!]
\centering
\includegraphics[scale=0.2]{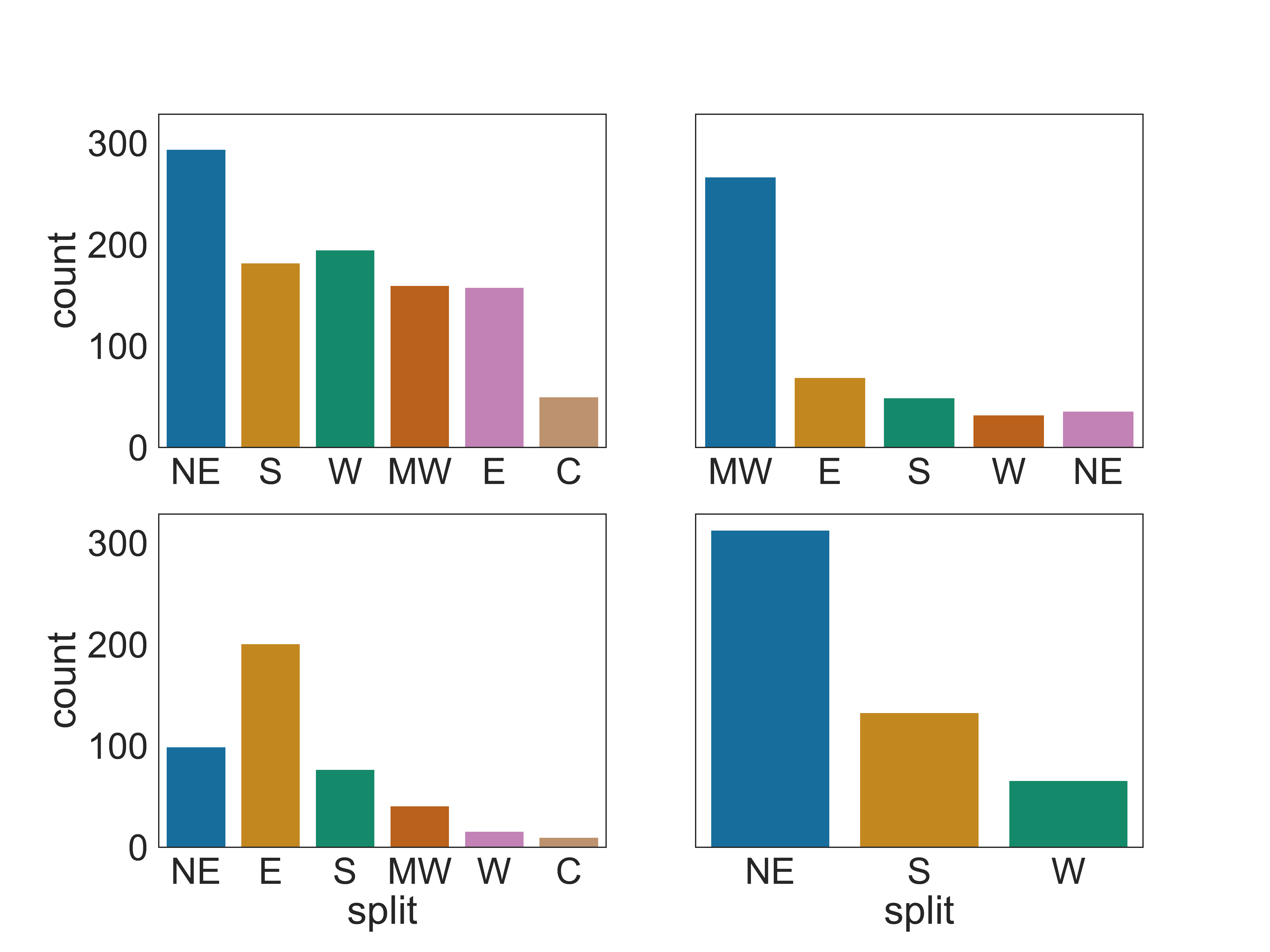}
\includegraphics[scale=0.2]{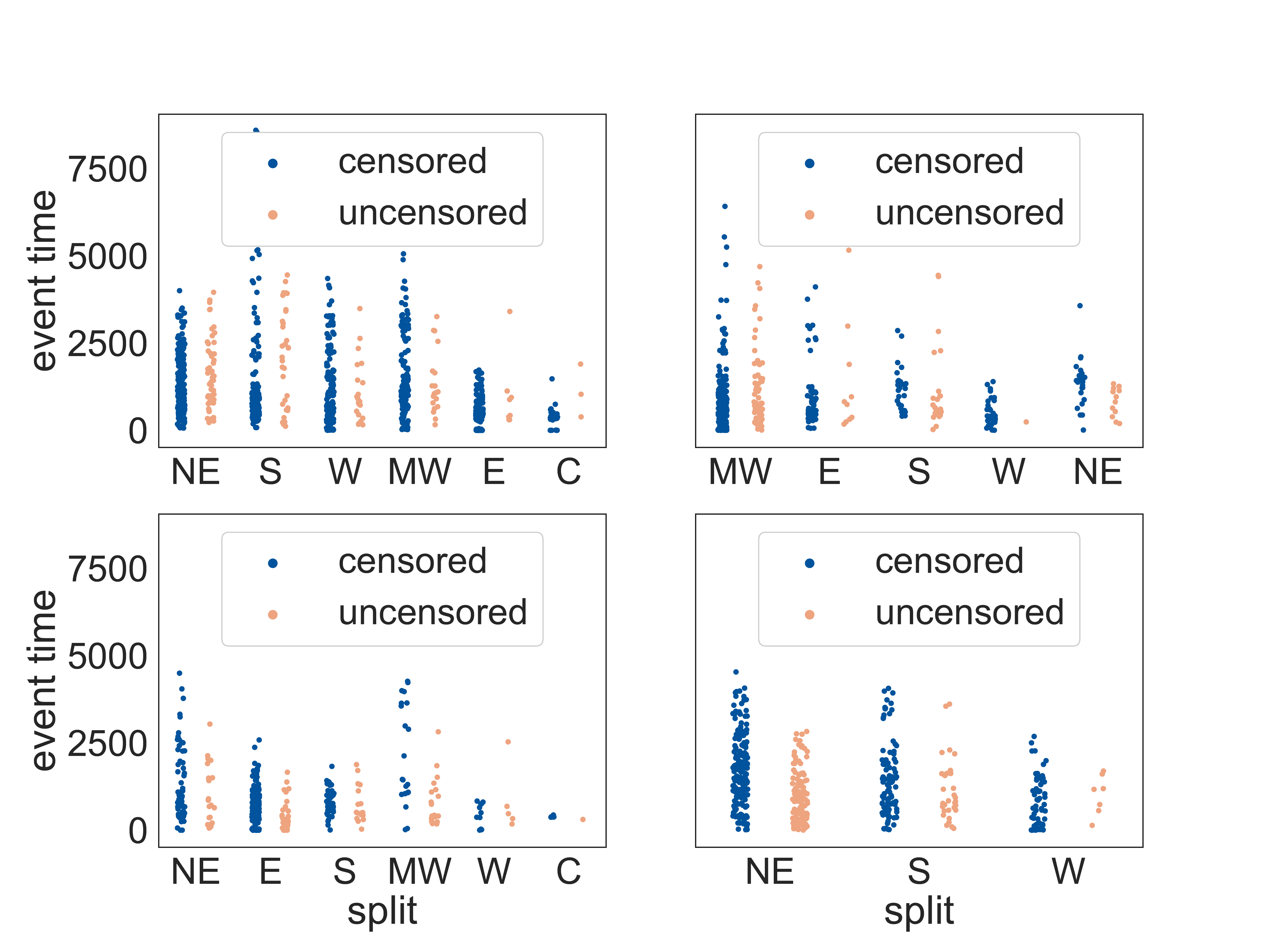}
\caption{Details on the histopathology datasets from TCGA. In each figure, from left to right and top to bottom, the cancer types are BRCA, LGG, COAD and KIRC. \textit{Left:} Number of patients per center. \textit{Right:} Distribution of event times per center. NE, S, W, MW, E, C respectively stand for Northeast, South, West, Middlewest, Europe and Canada.}
\label{fig:tcga_details_slides}
\end{figure}

Figs.~\ref{fig:tcga_details_tabular} and \ref{fig:tcga_details_slides} show the details of the data distributions for tabular and image data, respectively.

\section{Experimental details}
\label{app:experimental_details_global}
In all experiments, c-indices of the proposed discrete-time method are computed on the
true times $t_i$ and not the binned ones, which are only used for training.
This ensures a fair comparison of the c-indices of all reported approaches.
Further, in all cases, hyperparameters were manually tuned, following a
limited number of cross-validation runs.

\subsection{Section 6.1}

For pooled, local and ensembled baselines, the \texttt{lifelines} class
\texttt{CoxPHFitter} is used for fitting a Cox model with all default
parameters, in particular with no regularization. The version of
\texttt{lifelines}~\cite{cameron_davidson_pilon_2020_lifelines} used is~\texttt{0.22.8}.

Both minibatch and naive FL experiments are implemented using PyTorch~\cite{pytorch2019neurips} (version \texttt{1.14.0}), with
the Cox loss optimization performed using Adam~\cite{kingma2014adam} with a learning rate of
$10^{-3}$. Batches are sampled with replacement; in total 5000 batches of
size 100 each are used.
For the naive FL experiments, we simulate the
existence of multiple clients by looping through centers and storing the
produced gradients; once all clients have been seen, these stored gradients
are be aggregated, and the aggregate used by the optimizer.

The exact same parameters are used to minimize the BCE loss in the case of
discrete-time FL. Discrete times are created using $Q = 1$, so that all
unique event times are considered.

\subsection{Section 6.2}
\label{app:exp_details_tabular}
For boosting, hyperparameters such as the learning rate (\texttt{eta}), the
maximum depth of a tree (\texttt{max\_depth}) and the subsample ratio
(\texttt{subsample}) are fixed to respectively $0.01$, $3$, $0.5$,
reasonable values working generally well on such tabular data. The optimal
number of boosting rounds is set using an inner 5-fold cross-validation on
the training set. The version of \texttt{xgboost}~\cite{chen2016xgboost} used is~\texttt{0.90}.

For naive FL, the same parameters as in Section 6.1 are used. For discrete-time FL,
samples are created using $Q = 30$ (i.e. binning events by month), and 1000
batches of size 5000 are used. Moreover, a weighted BCE loss is used such
that positive samples are given more importance, following the ratio
between negative and positive samples.

\subsection{Section 6.3}
\label{app:exp_details_slides}
\subsubsection{Slide preprocessing}
\paragraph{Tile extraction}
We first extract matter-bearing, non-overlapping tiles from each histopathology slide.
For this purpose, a single U-Net~\cite{unet} is trained to separate all individual pixels between foreground (containing matter) and background, using manually annotated tiles of histology images.
Using this U-Net, we extract $8000$ tiles per slide.

\paragraph{ResNet feature extraction}
Each extracted tile is processed through an ImageNet-pretrained ResNet-50 deep neural network,
stopping at the penultimate layer of the architecture (before getting logits), leading
to a vector of size $2048$ per tile.

\paragraph{Dimensionality reduction}
At this stage of preprocessing, each slide has been transformed into an array of size~$8000 \times 2048$, where~$8000$ is the number of tiles extracted from the slide and $2048$ the number of features per tile.
Due to this large dimensionality and the few number of samples, to reduce overfitting, we additionally perform a feature reduction operation.
For each cancer's dataset, we train a linear autoencoder on ResNet features of tiles extracted from this dataset, with bottleneck dimension $256$ and no bias.
This autoencoder is trained for the reconstruction of the original tiles' features under the Mean-Square Error loss.
This training is carried out using the Adam optimizer with standard hyperparameters and learning rate $10^{-3}$ on a subset of $100,000$ randomly-sampled tiles for $3$ epochs.

\paragraph{Final representation}
After passing each tile's ResNet features through the encoder part of the trained autoencoder, each individual slide is finally represented by an array of shape $8000 \times 256$, which concludes preprocessing.

\subsubsection{Network architectures and training}
We first describe settings common to the baseline and discrete-time FL approaches.

\paragraph{Cross-validation}
We perform 5 runs of 5-fold cross-validation, each initialized with a different seed.
In each CV, the validation sets are built from the local centers' validation set.
In both cases, cross validation is done with a patient split.
In particular, during evaluation, patients with multiple slides have the scores for each slide averaged, yielding a single score per patient.
In the training folds, all slides belonging to a center are treated as independent samples.

\begin{figure}[h]
\centering
\includegraphics[scale=0.5]{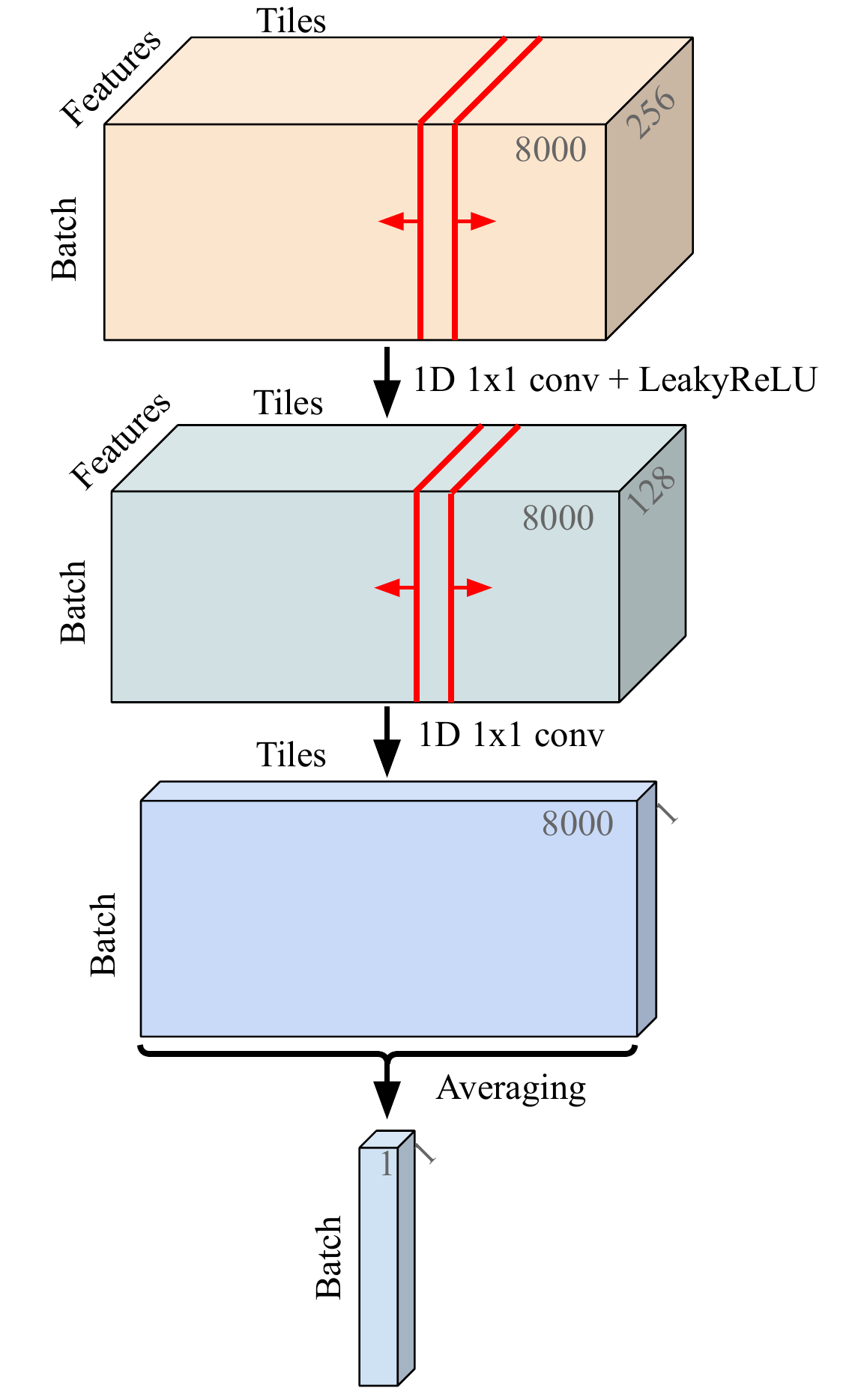}
\caption{Auxiliary model $\phi_{\theta}$ applied to preprocessed histopathology slide images.
The use of 1D $1 \times 1$ convolutions ensures that the same layer is applied to all tiles of a given slide.
The batch axis corresponds to multiple slides.
The negative slope of the LeakyReLU is set to $0.1$.
\label{fig:auxiliary_model}
}
\end{figure}

\paragraph{Auxiliary model}
The same auxiliary model,  \textit{i.e.} the function~$\bm{\phi}_\theta$, is the same in both approaches.
Its architecture is depicted in Figure~\ref{fig:auxiliary_model} and begins with a 1D $1 \times 1$ convolution of output size $128$, a
Leaky ReLU with negative slope of $0.1$, and another 1D $1 \times 1$ convolution of output
size 1.
Due to the use of $1 \times 1$ convolutions, each tile is assigned a single scalar score by a shared network.
Then, the average of all scores across tiles is computed, leading to a single score per slide, which is
the output of the auxiliary model.

\paragraph{Baseline training}
The baseline (MINI) experiments are carried out with Tensorflow~\cite{tensorflow2015-whitepaper} (version~\texttt{1.14}).
The network training uses the following hyperparameters: batch size $30$, $20$ epochs, optimizer Adam with learning rate set to $10^{-3}$ and standard hyperparameters otherwise.

\paragraph{Discrete-time FL training}
For discrete-time FL, experiments are carried with PyTorch (version \texttt{1.14.0}).
Batches of size 100 are sampled without
replacement, totalling 25 epochs.
The optimizer Adam is used, with learning rate set to $10^{-3}$ and standard hyperparameters otherwise.
Regarding the binning of times-to-event, once again, $Q$ is set to $30$ and
label balancing is used in the BCE loss.

\end{document}